\documentclass{article}
\usepackage{spconf,amsmath,graphicx}
\usepackage{xcolor}

\usepackage{mwe} 
\usepackage{url}
\usepackage{flushend}

\usepackage{fancyhdr}
\fancypagestyle{firstpage}{
  \fancyhf{}
  \fancyhead[C]{\vspace{-2.2cm}Extended version of the paper ``Information Flow Through U-Nets,'' to be presented at IEEE International Symposium on Biomedical Imaging (ISBI), Nice, France, April 13-16, 2021. \vspace{0.3cm}} 
}

\title{Analysis of Information Flow Through U-Nets}
\name{Suemin Lee and Ivan V. Baji\'{c}\thanks{This work was supported in part by the Natural Sciences and Engineering Council (NSERC) of Canada.}}
\address{School of Engineering Science, Simon Fraser University, Canada}

\begin{document}
%
\maketitle
\begin{abstract}
Deep Neural Networks (DNNs) have become ubiquitous in medical image processing and analysis. Among them, U-Nets are very popular in various image segmentation tasks. Yet, little is known about how information flows through these networks and whether they are indeed properly designed for the tasks they are being proposed for. In this paper, we employ information-theoretic tools in order to gain insight into information flow through U-Nets. In particular, we show how mutual information between input/output and an intermediate layer can be a useful tool to understand information flow through various portions of a U-Net, assess its architectural efficiency, and even propose more efficient designs.  

\end{abstract}
\thispagestyle{firstpage}

\section{Introduction}
\label{sec:intro}


Deep Neural Networks (DNNs) have become common solutions for a variety of computer vision problems and have been particularly popular in medical image processing and analysis. Among the various DNN architectures, U-Nets~\cite{Olaf} are considered a gold-standard in image segmentation, not just for medical images, but also in more general settings. Yet, little is known about information flow in these networks. There are few strategies, besides trial-and-error, to assess the effectiveness of a particular U-Net design. 

In this paper, we use mutual information~\cite{Cover} to study information flow through U-Nets, which helps reveal the inner workings of these networks. Inspired by the information bottleneck theory~\cite{tishby_information_1999}, mutual information has recently become a popular tool for studying deep learning models~\cite{tishby_deep_2015,shwartz-ziv_opening_2017}. The power of mutual information comes from the fact that it can quantify arbitrary -- linear or nonlinear -- dependence between random quantities, yet has a simple interpretation in terms of bits of information. Existing works on the measurement of information flow through deep models have focused on single-stream DNNs~\cite{tishby_deep_2015, shwartz-ziv_opening_2017,Andrew,strouse_deterministic_2016,kolchinsky_caveats_2018, goldfeld_2019}. To our knowledge, this paper is the first to perform and interpret such measurements in a multi-stream DNN (specifically, U-Net), and the first to consider a DNN model made for segmentation.     

In Section~\ref{sec:preliminaries}, we introduce the necessary concepts from information theory, specifically the mutual information and the data processing inequality. Section~\ref{sec:methods} describes models, data, and the specifics of estimating mutual information for an image segmentation task. Experiments and analysis are presented in Section~\ref{sec:experiments}, including a novel concept of U-Plot, a useful tool for understanding information flow in U-Nets. The paper is concluded in Section~\ref{sec:conclusions}. Code is available on GitHub.\footnote{\url{https://github.com/Suemin-Lee/MI_Unet}}

\section{Preliminaries}
\label{sec:preliminaries}

\subsection{Mutual information}
Consider two discrete quantities $X$ and $Y$ defined over discrete sample spaces $\mathcal{X}$ and $\mathcal{Y}$, respectively. In our case, $X$ and $Y$ will be tensors. Lowercase letters ($x,y$) will denote their specific realizations. Let $p(x,y)$ be the joint distribution of $X$ and $Y$. The mutual information (MI) between $X$ and $Y$, denoted $I(X;Y)$, is defined as
\begin{equation}
  I(X;Y)= \sum_{x,y}  p(x,y)\,\log_2 \frac{p(x,y)}{p(x)p(y)} \ .
  \label{eq:mutual_information}
\end{equation}
With $\log_2$ in the definition, the units of MI are bits. Other bases of the logarithm lead to other units (nats, Hartleys, etc.). MI is symmetric ($I(X;Y)=I(Y;X)$) and measures how different is the joint distribution $p(x,y)$ from the product of the marginal distributions $p(x)p(y)$. If $X$ and $Y$ are independent, $I(X;Y)=0$, otherwise $I(X;Y)>0$. By recognizing that $p(x,x)=p(x)$, it is easy to see that $I(X;X)=  -\sum_{x} p(x) \log_2 p(x) = H(X)$, the entropy~\cite{Cover} of $X$.

\subsection{Estimating mutual information}
\label{sec:estimating_MI}
In order to estimate MI from empirical measurements, one needs to be able to estimate $p(x,y)$. From there,  marginals $p(x)$ and $p(y)$ can be computed and finally $I(X;Y)$ can be obtained from~(\ref{eq:mutual_information}). While the procedure seems straightforward, the difficulty lies in the fact that the domain of $p(x,y)$, $\mathcal{X}\times \mathcal{Y}$, is usually very high-dimensional, yet the number of data points from which $p(x,y)$ needs to be estimated is comparatively very small.

A number of methods have been proposed for estimating MI~\cite{Andrew,Kolchinsky1,belghazi_mutual_2018}. In this work, we utilize the two methods presented by ~\cite{Andrew}. One of these is a histogram-based method, where the space $\mathcal{X}\times \mathcal{Y}$ is partitioned into bins, and frequency counts in histogram bins are used as estimates of $p(x,y)$. The other method is a kernel density estimator (KDE)~\cite{Andrew}, employing a Gaussian kernel to obtain estimates of $p(x,y)$.

\subsection{Data processing inequality}
\label{sec:dpi}
For a Markov chain $X\to M \to Y$, data processing inequality (DPI)~\cite{Cover} states that
\begin{equation}
    I(X;M) \geq I(X;Y).
\end{equation}
It has been recognized~\cite{tishby_deep_2015,shwartz-ziv_opening_2017} that single-stream DNNs behave as Markov chains. Hence, a single-stream DNN with input $X$, output $Y$, and $L$ hidden layers ($M_1,M_2,...,M_L$), can be represented as a Markov chain $X \to M_1 \to M_2 \to ... \to M_L \to Y$, from which it follows that
\begin{equation}
    I(X;M_1)\geq I(X;M_2) \geq ... \geq I(X;Y).
    \label{eq:DPI_DNN}
\end{equation}
By recognizing that Markovity also holds in reverse~\cite{Cover}, we can also conclude that: 
\begin{equation}
    I(X;Y) \leq I(M_1;Y) \leq I(M_2;Y) \leq ... \leq I(M_L;Y).
    \label{eq:DPI_DNN_reverse}
\end{equation}

It should be noted that U-Nets are not single-stream models, due to skip connections that carry data over from earlier layers. Hence, it is to be expected that some of the inequalities in~(\ref{eq:DPI_DNN})-(\ref{eq:DPI_DNN_reverse}) will not hold in a U-Net. Indeed, experimental results will confirm this.

\section{Methods}
\label{sec:methods}

\subsection{Models and data}

\begin{figure}[!ht]
\begin{center}
\includegraphics[width=0.47\textwidth]{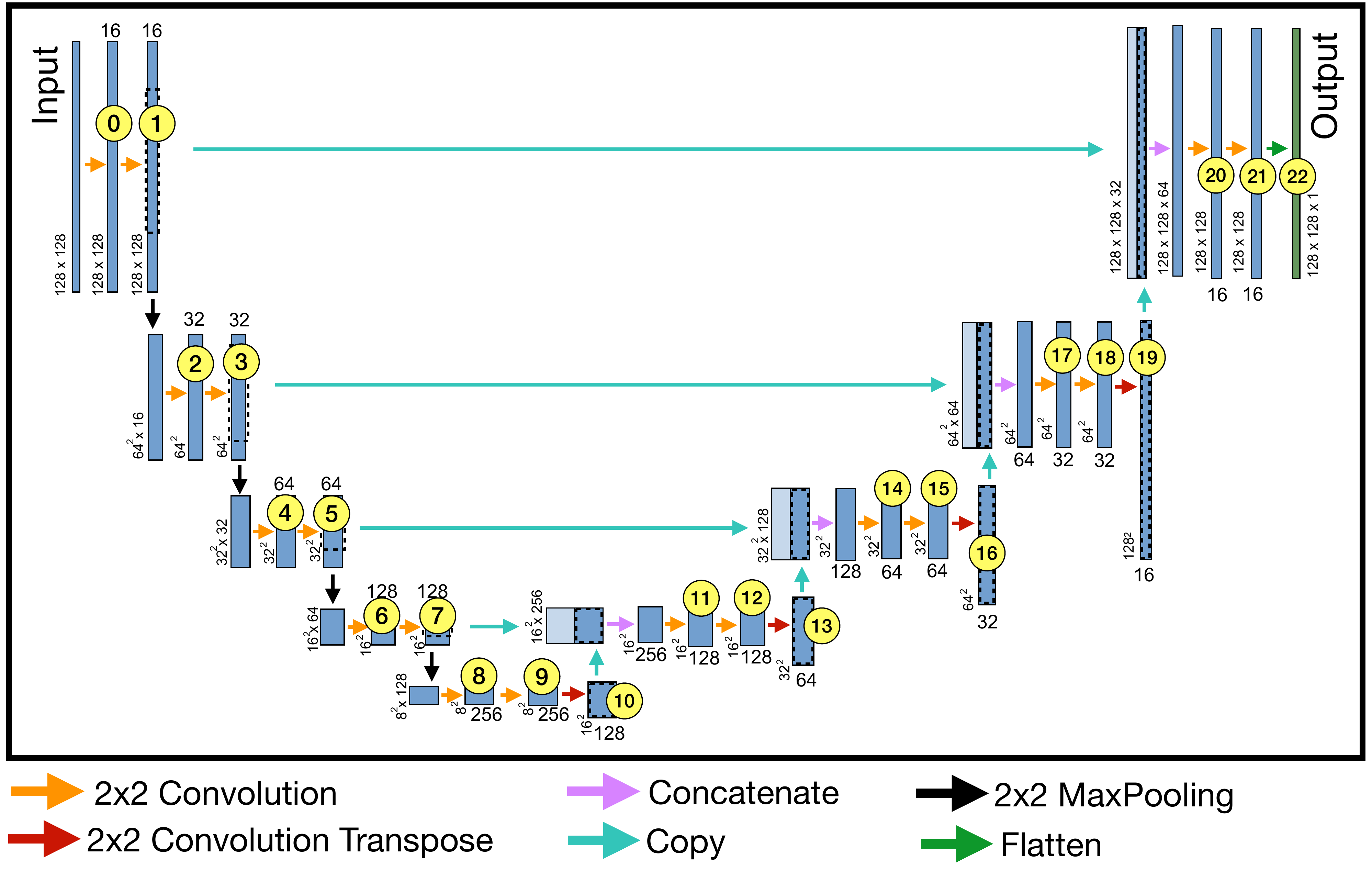}
\caption {Original U-Net model architecture modified to handle $128 \times 128$ input images. Numbers in yellow circles indicate layer index.
} \label{fig:U_net_model}
\end{center}
\end{figure}
Following the common practice in information-theoretic analysis of DNNs~\cite{tishby_deep_2015, shwartz-ziv_opening_2017,Andrew,strouse_deterministic_2016,kolchinsky_caveats_2018, goldfeld_2019}, in the experiments we focused a smaller, canonical model and its variations, rather than large state-of-the-art models. The primary reason is that in a complex model, many factors may influence the results, and it becomes increasingly more difficult to disentangle various side-effects from the main issues being studied.

The U-Net model employed in our experiments, shown in Fig.~\ref{fig:U_net_model}, follows the original architecture proposed in~\cite{Olaf}, modified to handle $128 \times 128$ input images. This model and its few variations were trained and tested on the cell image dataset from~\cite{Juan}. This dataset contains 670 segmented nuclei images.  The images were acquired under various conditions, and various cell types are included as samples with different magnifications and imaging modality (bright field vs. fluorescence). Of the 670 images, 605 were used for training, and 65 were used for validation/testing.

\subsection{Estimating MI for segmentation masks}

The DNNs studied in~\cite{tishby_deep_2015, shwartz-ziv_opening_2017,Andrew,strouse_deterministic_2016,kolchinsky_caveats_2018, goldfeld_2019} were all classifiers with a relatively small number of classes, where 
it is relatively easy to obtain estimates of $I(M_i;Y)$ following the approach in~\cite{Andrew}. However, the  U-Net in Fig.~\ref{fig:U_net_model} produces binary segmentation masks of resolution $128 \times 128$. This U-Net can also be considered as a classifier, but the number of classes is now  $2^{128\cdot 128}=2^{16,384}\approx 10^{4,932}$, making it hard to estimate $I(M_i;Y)$ reliably on a relatively small set of images. For this reason, we employed two methods for output dimensionality reduction -- spatial coarsening and K-means clustering -- in order to estimate $I(M_i;Y)$. It should be noted that these dimensionality reduction methods are used only to estimate $I(M_i;Y)$, while the model is still trained to produce $128 \times 128$ segmentation masks. 

\textbf{Spatial coarsening.} In this approach, the resolution of the output segmentation mask is reduced by dividing it into coarser blocks. Specifically, the $128\times 128$ segmentation mask is divided into $16$ blocks of size $32\times 32$. The value of the block is set to $1$ if the number of pixels within the blocks that are equal to $1$ exceeds the threshold $T$, otherwise the value of the block is set to $0$. For the experiments, we chose $T=64$. 

Since there are $16$ blocks in the coarsened segmentation masks, the number of possible outputs is now $2^{16} = 65,536$. The maximum output entropy $H(Y)$ in this case is $\log_{2}65,536=16$ bits, achieved when all classes are equally likely~\cite{Cover}. Since $M_L \to Y \to Y$ is a Markov chain, we have from DPI that $I(M_L;Y) \leq I(Y;Y) = H(Y)$, and combining this with~(\ref{eq:DPI_DNN_reverse}), we conclude that $I(M_i;Y)\leq 16$ bits in this case. Had we chosen a different number of blocks per mask (other than $16$), we would have obtained a different upper bound on $I(M_i;Y)$. Hence, estimates of $I(M_i;Y)$ in this case are to be taken as indicators of the relative amount of information that different layers $M_i$ carry about the output $Y$, rather than in absolute terms. 

\textbf{K-means clustering.} Another way to reduce the number of output classes is K-means clustering~\cite{duda_et_al_2000}. We vectorized the output segmentation mask and employed K-means clustering with $K = 64$ clusters. This leads to the maximum output entropy of $H(Y)=\log_2 64 = 6$ bits, achieved when all clusters are equally likely, which is an upper bound on all $I(M_i;Y)$. With another value of $K$, we would have another upper bound $\log_2K$. As with spatial coarsening, the numerical values of estimated $I(M_i;Y)$ should be taken as indicators of the relative amount of information that different layers $M_i$ carry about the output $Y$, rather than in absolute terms. 

\begin{figure}[htb]

\begin{minipage}[b]{1.0\linewidth}
  \centering
  \centerline{\includegraphics[width=8.5cm]{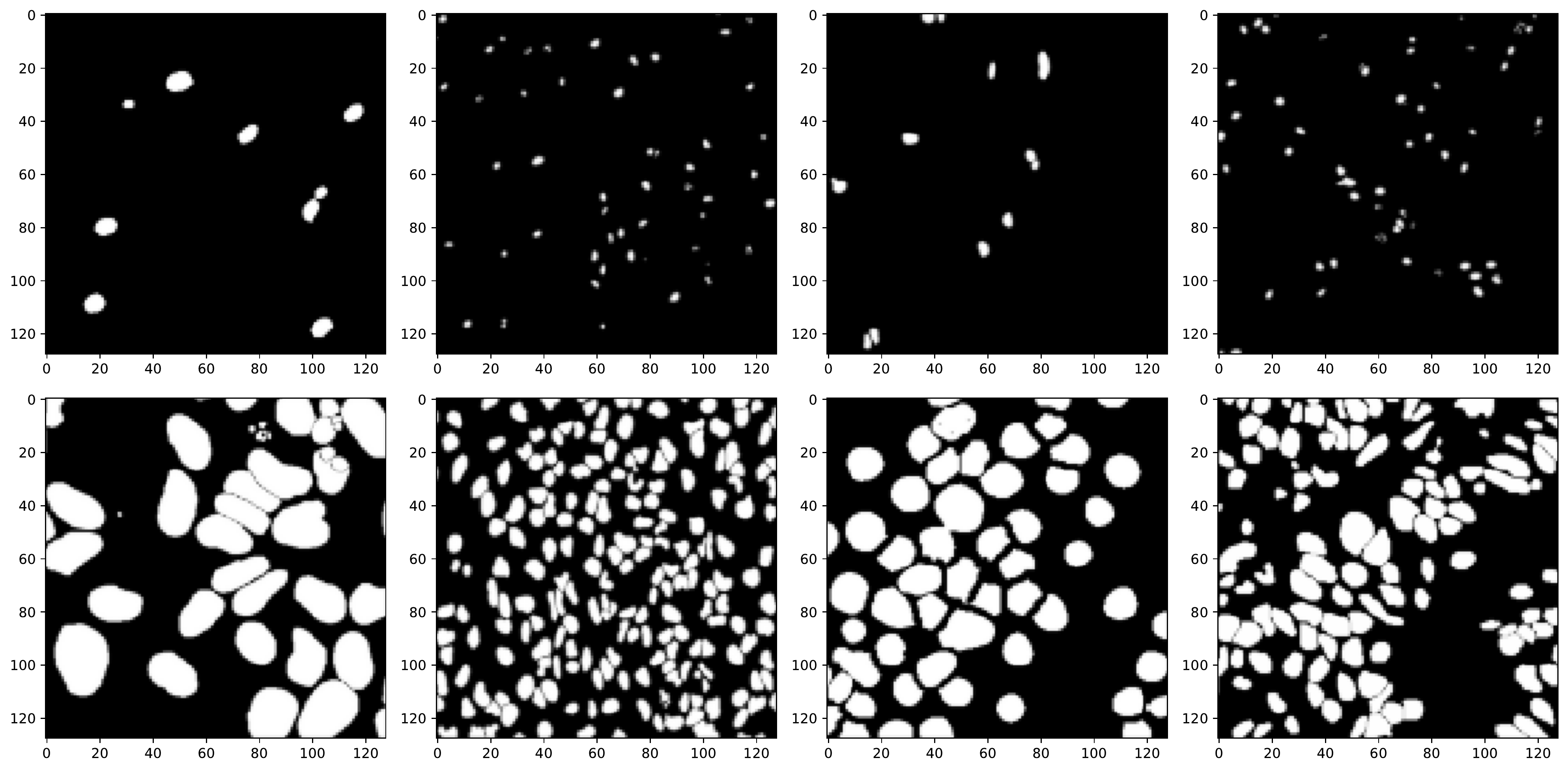}}
  \centerline{(a) Clusters created by spatial coarsening}\medskip
\end{minipage}

\begin{minipage}[b]{1.0\linewidth}
  \centering
  \centerline{\includegraphics[width=8.5cm]{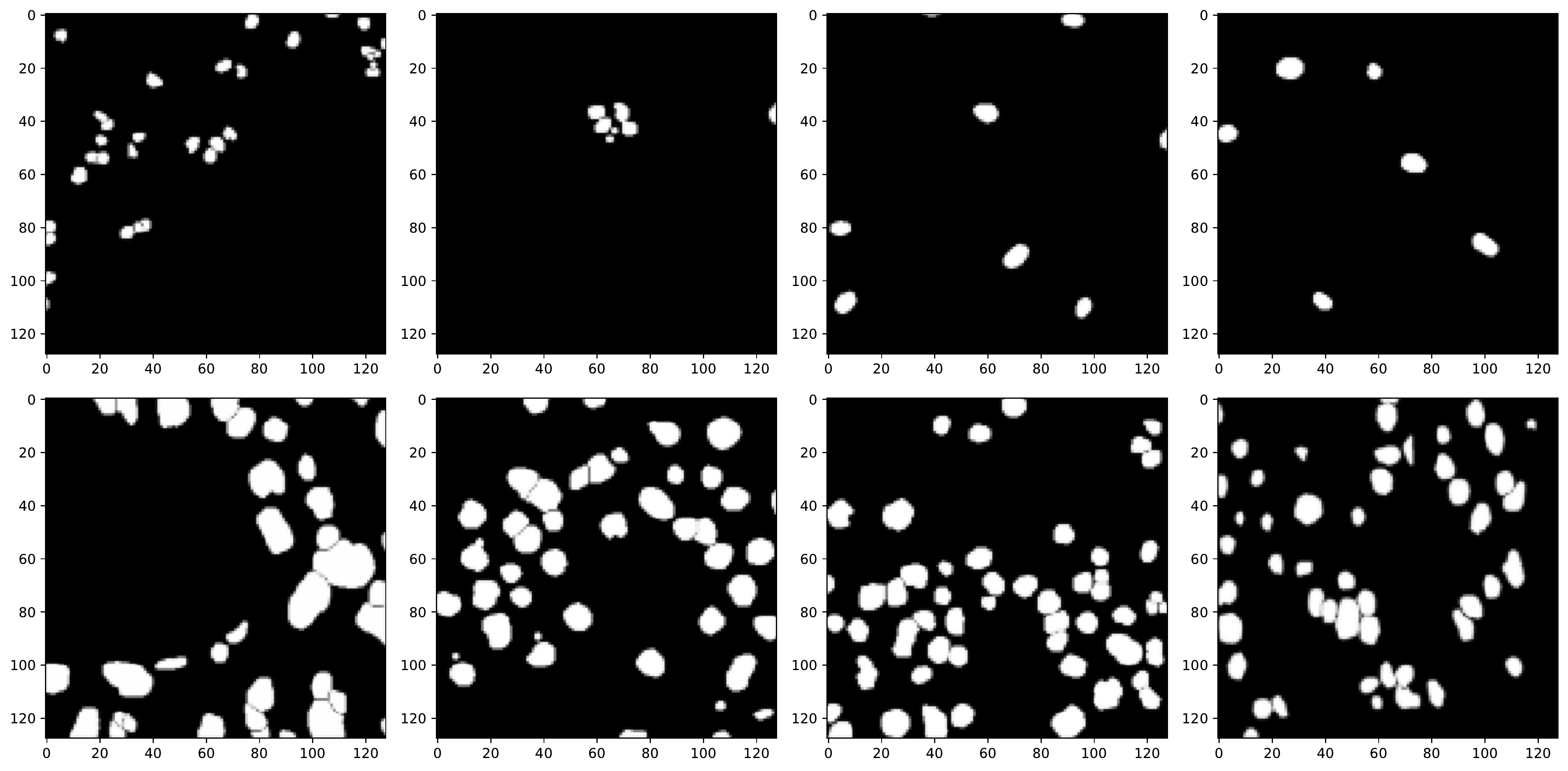}}
  \centerline{(b) Clusters created by K-means}\medskip
\end{minipage}
\caption{Each row shows four segmentation masks from the same cluster. Clusters are created by (a) spatial coarsening and (b) K-means clustering.}
\label{fig:Mask_image}
\end{figure}

Fig.~\ref{fig:Mask_image} shows examples of segmentation masks clustered by spatial coarsening and K-means, where each row shows four examples from one cluster. Spatial coarsening (Fig.~\ref{fig:Mask_image}(a)) creates clusters that have similar density in various image blocks, but is agnostic to the actual morphology of the segments. On the other hand, K-means clustering (Fig.~\ref{fig:Mask_image}(b)) creates more morphologically meaningful clusters. For this reason, in our experiments we used K-means clustering.

\section{Experiments and Analysis}
\label{sec:experiments}

\subsection{Information plane analysis}
\begin{figure*}[ht]
\centering
    \includegraphics[width=1\textwidth]{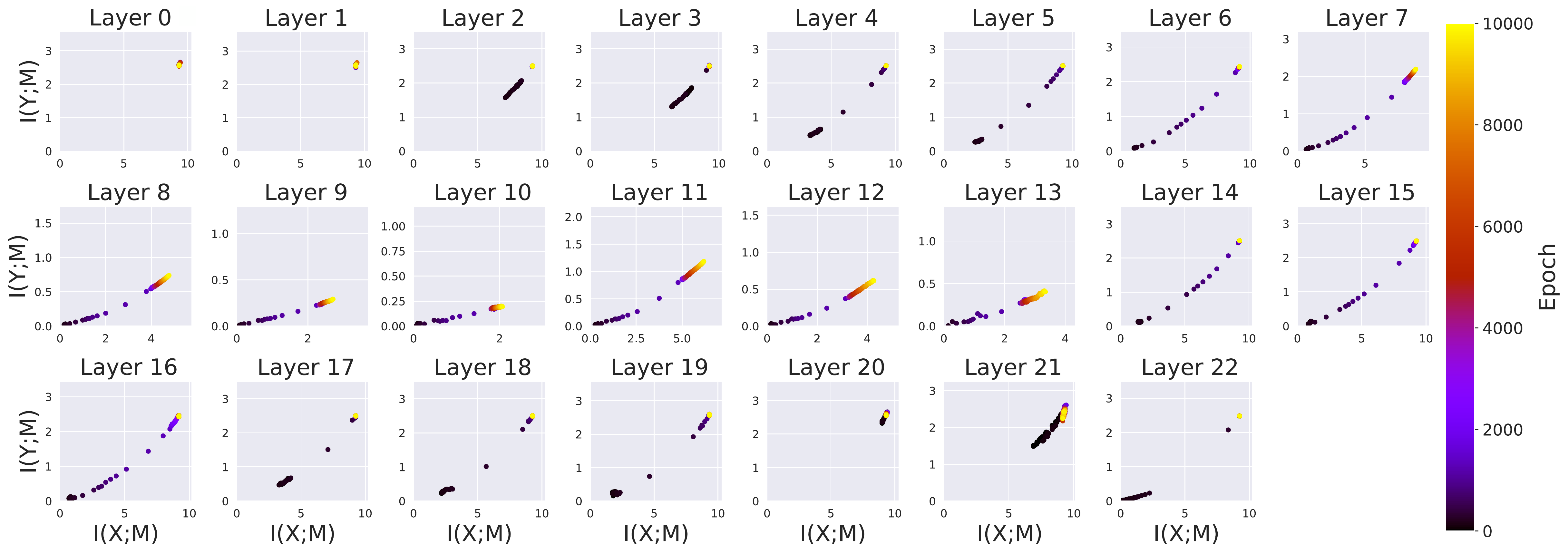}
    \caption{Information planes of the 23 layers of U-Net from Fig.~\ref{fig:U_net_model}. Mutual information was estimated using the KDE estimator~\cite{Andrew} with noise variance 1, for a K-means clustered output with $K=64$. }
\label{fig:Break_down_infoplane}
\end{figure*}

\emph{Information plane}~\cite{tishby_deep_2015,shwartz-ziv_opening_2017} is a plane where the x-axis is $I(X;M_i)$, the mutual information between the input $X$ and the $i$-th layer $M_i$, while the y-axis is $I(Y;M_i)$, the mutual information between the output $Y$ and the $i$-th layer $M_i$. Observing the dynamics of mutual information in this plane helps us gain a better understanding of how a model learns.  Fig.~\ref{fig:Break_down_infoplane} shows the information planes of the 23 layers of the U-Net from Fig.~\ref{fig:U_net_model}, obtained over 10,000 training epochs. Mutual information was estimated using the KDE method~\cite{Andrew} with unit noise variance, and all other parameters left as default. Similar results were obtained using the histogram-based estimator~\cite{Andrew} with a bin size of 0.2, but those plots are not shown since they are somewhat redundant. For the purpose of computing $I(Y;M_i)$, the output was clustered using K-means clustering with $K = 64$. Spatial coarsening led to similarly-shaped information curves, but with numerically different values, since the probability distributions under spatial coarsening are estimated over different spaces and lead to different upper bounds on mutual information, as discussed earlier. 

The first thing to note about information plots in Fig.~\ref{fig:Break_down_infoplane} is that $I(Y;M_i)$ goes up to about 3 bits for some of the layers, and less for others. Since $K=64$ for the clustered output, the upper bound on $I(Y;M_i)$ is $\log_2 64 = 6$ bits, but this would only be achieved if all clusters were equally likely. Evidently, this is not the case, and mutual information is less than 6 bits. 

Perhaps the main observation from Fig.~\ref{fig:Break_down_infoplane} is that different layers in a U-Net learn at different rates. Mutual information measured at different epochs is color-coded according to the colormap shown on the right. Earlier epochs are darker, and latter are lighter, finishing with yellow at epoch 10,000. The first two layers learn very quickly - within the first few epochs, $I(Y;M_i)$ reaches its maximum around 3 bits, and stays there throughout the training. This is why only the yellow dot is visible in these plots. Meanwhile, other layers take longer to reach their maximum mutual information with respect to the output. For example, layers 8--14, which are at the bottom of the U-Net, all seem to still be learning at epochs 6,000--8,000, since their information curves show some red and orange points. Also, these layers at the bottom of the U-Net do not reach $I(Y;M_i)$ of 3 bits, suggesting that they contain less information about the output than some of the other layers. We examine this observation in more detail in the next section.   

\subsection{U-Plot analysis}

\begin{figure*}[h]
\centering
\includegraphics[width=0.8\textwidth]{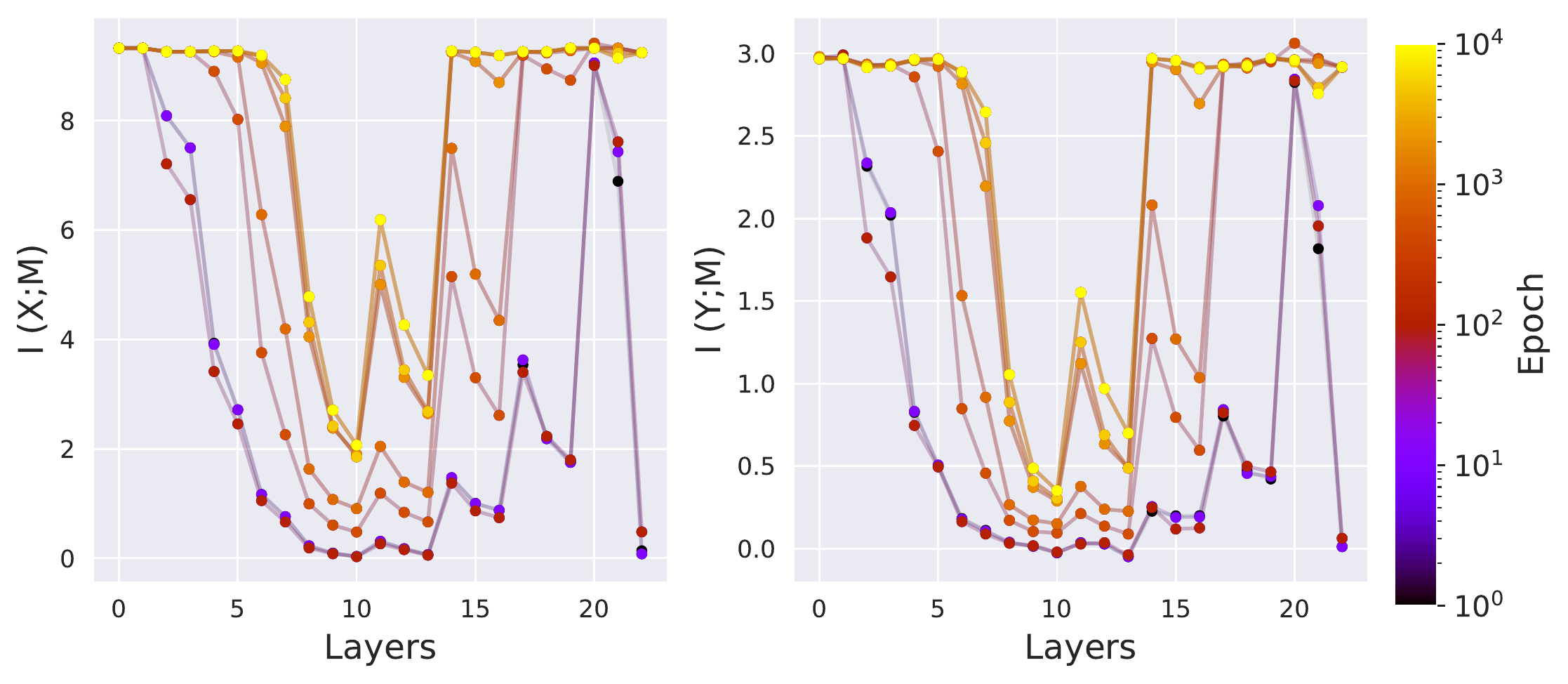}
 \caption {U-Plots of $I(X;M_i)$ (left) and $I(Y;M_i)$ (right). } 
\label{fig:U_shaped_KDE_MI}
\end{figure*}

In order to gain better insight into learning within a U-Net, we plot $I(X;M_i)$ vs. layer index $i$ for all layers in 
Fig.~\ref{fig:U_shaped_KDE_MI}. Another plot is created for $I(Y;M_i)$ vs. $i$. Mutual information (color-coded) is shown at various training epochs, and values obtained at the same epoch at different layers are connected by solid lines. We refer to these plots as \emph{U-Plots} because they exhibit a U-shape, similar to the U-Net itself.   

The reason for the U-shape in the U-Plots is as follows. Based on the data processing inequality~(\ref{eq:DPI_DNN}), $I(X;M_i)$ is a decreasing function of $i$ for a single-stream model. However, U-Net is a multi-stream model, where data is carried via skip connections from earlier to latter layers: $M_1 \to M_{20}$, $M_3 \to M_{17}$, $M_5 \to M_{14}$, and $M_7 \to M_{11}$. At layers 11, 14, 17, and 20, two data streams merge, so Markovity gets violated and data processing inequality does not hold. Indeed, one can see in Fig.~\ref{fig:U_shaped_KDE_MI} that $I(X;M_i)$ goes up at these merge layers in early iterations, and the same happens to $I(Y;M_i)$. 

Next, we examine more carefully what happens at merge layers. As the training goes on, mutual information increases, as we already saw in Fig.~\ref{fig:Break_down_infoplane}. $I(Y;M_{20})$ reaches the maximum of about 3 bits with around 1,000 epochs, and so does $I(Y;M_{17})$. However, $I(Y;M_{14})$ clearly needs more iterations to reach the maximum, since we clearly see an orange-colored point at $I(Y;M_{14}) \approx 2$ bits, meaning that the maximum mutual information of around 3 bits has not been reached in 1,000 iterations. Eventually, $I(Y;M_{14})$ does reach the maximum. However, $I(Y;M_{11})$ does not get anywhere near the maximum within 10,000 iterations and only reaches up to about 1.5 bits.

What can we conclude from the above analysis? Since $I(Y;M_{17})$ reaches the maximum with around 1,000 epochs, it means that at this point, the skip connection $M_1 \to M_{20}$ no longer carries any useful information. Recall that the role of merge layers is to bring mutual information up. But if layer 17 has already reached the maximum information about the output, then no useful information will be added at the merge point at layer 20. In other words, the skip connection $M_1 \to M_{20}$ should no longer be needed. Similarly, $I(Y;M_{14})$ reaches the maximum somewhere between iteration 1,000 and iteration 10,000. It is not quite clear when, because in Fig.~\ref{fig:U_shaped_KDE_MI}  iterations are shown exponentially spaced, but somewhere in that range $I(Y;M_{14})$ reaches the maximum, which means that at that point, skip connections $M_1 \to M_{20}$ and $M_3 \to M_{17}$ will no longer be needed. Lastly, $I(Y;M_{11})$ does not reach the maximum within 10,000 epochs, so the skip connection $M_5 \to M_{14}$ is required to allow the network to reach the best performance. 

We now put this reasoning to the test. If the assertions made above are true, then a U-Net with the skip connection $M_1 \to M_{20}$ removed should be able to perform just as well as the original U-Net, if allowed to train for around 1,000 epochs. Similarly, a U-Net with $M_1 \to M_{20}$ and $M_3 \to M_{17}$ removed should be able to reach the original U-Net's performance with training between 1,000 and 10,000 epochs. However, a U-Net with $M_1 \to M_{20}$, $M_3 \to M_{17}$, and $M_5 \to M_{14}$ removed would not be able to reach the performance of the original U-Net in 10,000 epochs. 

\begin{figure}[ht]
\begin{center}
\includegraphics[width=0.45\textwidth]{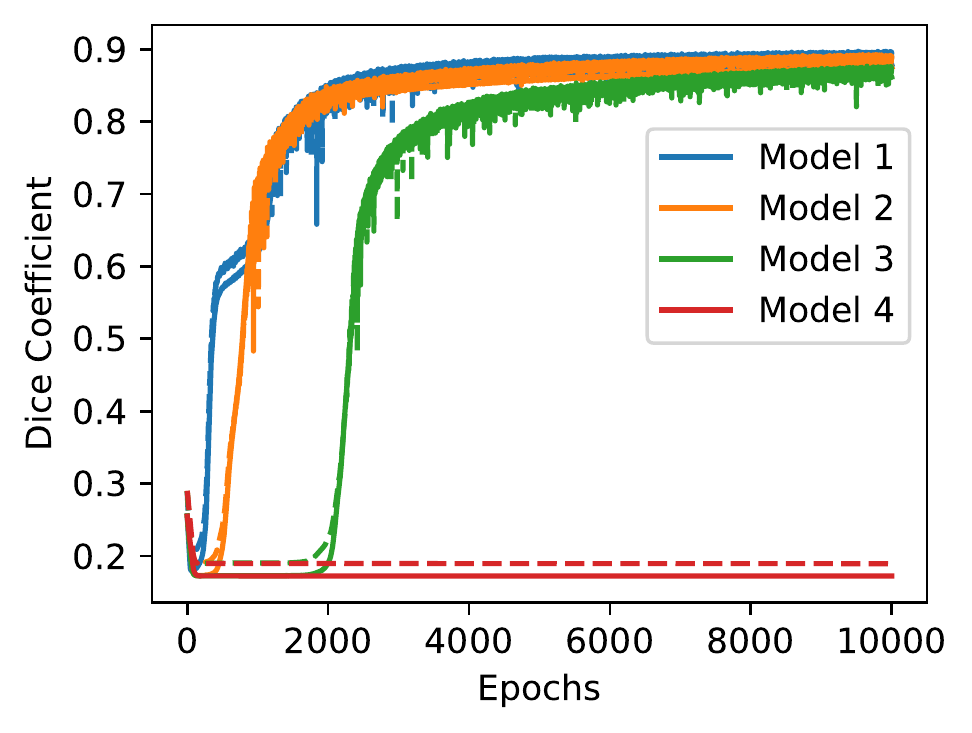}
\caption {Dice coefficient on the training set (solid) and validation set (dashed) for four models. 
} \label{fig:Accuracy_plot}
\end{center}
\end{figure}

To verify these assertions, we performed the following experiment. We trained four U-Nets for 10,000 epochs: \vspace{-5pt}
\begin{itemize}
    \item Model 1: original U-Net (Fig.~\ref{fig:U_net_model}) \vspace{-5pt}
    \item Model 2: U-Net with the top skip connection ($M_1 \to M_{20}$) removed \vspace{-5pt}
    \item Model 3: U-Net with the top two skip connections ($M_1 \to M_{20}$ and $M_3 \to M_{17}$) removed \vspace{-5pt}
    \item Model 4: U-Net with the top three skip connections ($M_1 \to M_{20}$, $M_3 \to M_{17}$, and $M_5 \to M_{14}$) removed \vspace{-5pt}
\end{itemize}
Their training and validation accuracy was measured in terms of Dice coefficient during the training, and the results are shown in Fig.~\ref{fig:Accuracy_plot}. From the figure, we see that Models 1 and 2 reach the maximum accuracy somewhere between epoch 2,000 and 3,000. Model 3 reaches the maximum accuracy after about 8,000 epochs, while Model 4 never reaches the maximum accuracy. All these results confirm the predictions made earlier: skip connection $M_1 \to M_{20}$ is not needed if the training is longer than 2,000 epochs; skip connections $M_1 \to M_{20}$ and $M_3 \to M_{17}$ are not needed if one is willing to train even longer (but less than 10,000 epochs); and skip connection $M_5 \to M_{14}$ cannot be removed if maximum performance is to be achieved within 10,000 epochs.

\begin{figure}[ht]
\begin{center}
\includegraphics[width=0.45\textwidth]{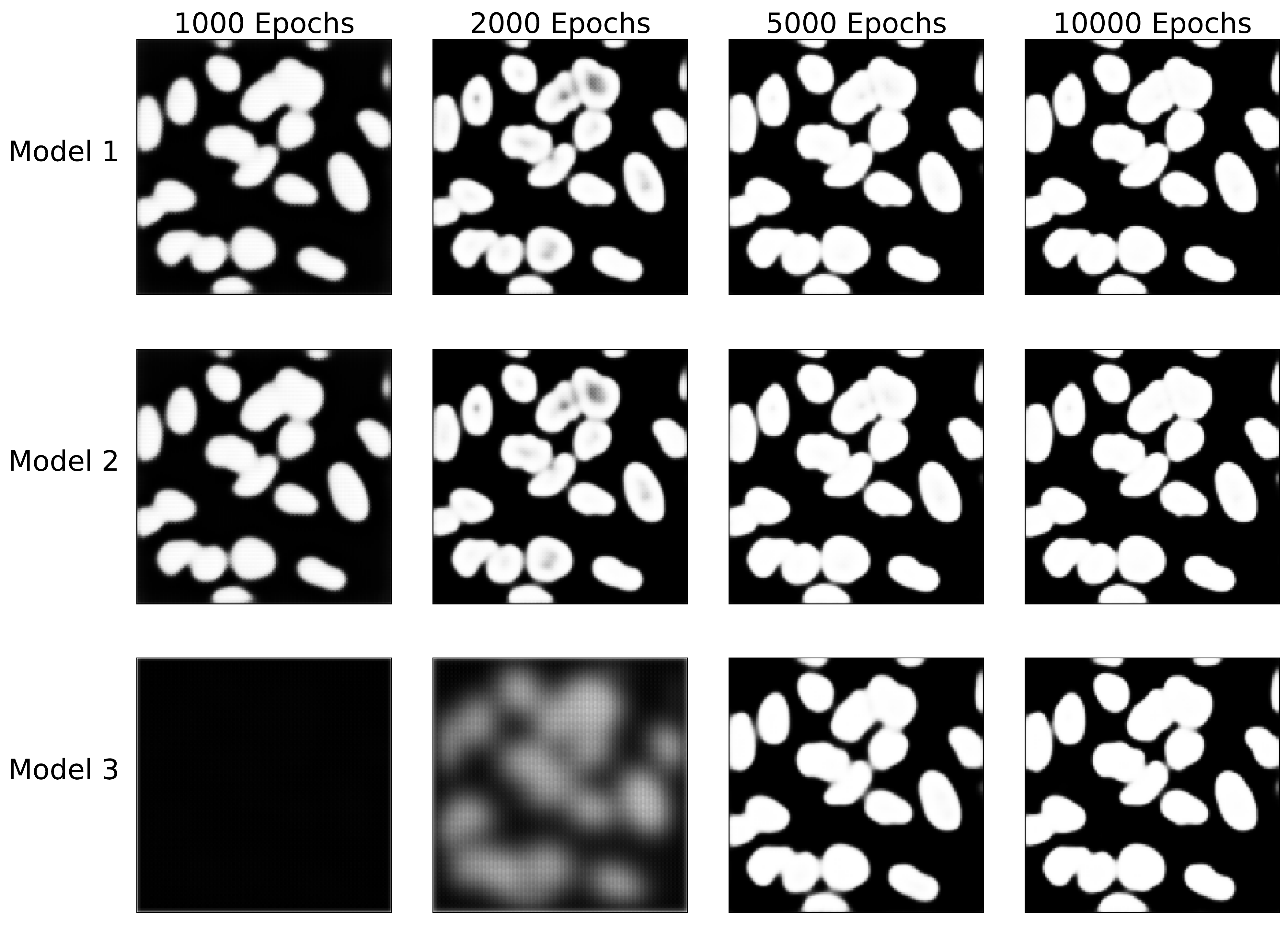}
\caption {
Examples of segmentation masks produced by Models 1-3 at different epochs (1,000, 2,000, 5,000, 10,000). 
} 
\label{fig:Segmented_mask_images}
\end{center}
\end{figure}

Fig.~\ref{fig:Segmented_mask_images} shows examples of segmentation masks produced by Model 1 (original U-Net), and Models 2 and 3, which are simplified versions of Model 1 with some skip connections removed, as described earlier. At 1,000 epochs, Models 1 and 2 produce reasonable segmentation masks, while Model 3 has not yet learned to produce segmentation masks. As training goes on, Models 1 and 2 improve slightly, while Model 3 catches up with them somewhere between 5,000 and 10,000 epochs. After 10,000 epochs all three models produces segmentation masks of similar quality. In particular, Models 2 and 3 produce very similar segmentation masks to those produced by Model 1, although they are architecturally simpler. This agrees with earlier analysis based on U-Plots in Fig.~\ref{fig:U_shaped_KDE_MI} as well as Dice coefficient results in Fig.~\ref{fig:Accuracy_plot}.

\section{Conclusions}
\label{sec:conclusions}
In this paper, we analyzed information flow through U-Nets using mutual information and introduced U-Plots, which have been shown to be useful for understanding how U-Nets learn. Based on U-Plot analysis, we made predictions about how the U-Net architecture could be modified and what would be the effect, and these predictions were verified experimentally. It was shown that, depending on the training duration, not all skip connections in a U-Net are necessary. This illustrates that the effectiveness of a particular network architecture is not absolute, but depends on how long the network is trained. More generally, we expect that the methodology presented here will contribute to a better understanding of multi-stream neural networks and give guidance for their more principled design.



\section{Compliance with Ethical Standards}
\label{sec:ethics}

This research study was conducted using data made available in open access. Ethical approval was not required with the open-access data.

\section{Acknowledgments}
\label{sec:acknowledgments}
This work was funded by the Natural Sciences and Engineering Research Council of Canada (NSERC). Computational resources were provided by Compute Canada.

\bibliographystyle{IEEEbib}
\bibliography{strings,refs}

\begin{thebibliography}{10}

\bibitem{Olaf}
O.~Ronneberger, P.~Fischer, and T.~Brox,
\newblock ``{U-Net:} convolutional networks for biomedical image
  segmentation,''
\newblock in {\em Proc. MICCAI}, 2015, pp. 234--241.

\bibitem{Cover}
T.~M. Cover and J.~A. Thomas,
\newblock {\em Elements of Information Theory},
\newblock Wiley, 2nd edition, 2006.

\bibitem{tishby_information_1999}
N.~Tishby, F.~C. Pereira, and W.~Bialek,
\newblock ``The information bottleneck method,''
\newblock in {\em Proc. Allerton Conference on Communication, Control, and
  Computing}, Monticcllo, Illinois, 1999, pp. 368--377.

\bibitem{tishby_deep_2015}
N.~Tishby and N.~Zaslavsky,
\newblock ``Deep learning and the information bottleneck principle,''
\newblock in {\em Proc. IEEE {Information} {Theory} {Workshop} ({ITW})}, 2015.

\bibitem{shwartz-ziv_opening_2017}
R.~Shwartz-Ziv and N.~Tishby,
\newblock ``Opening the black box of deep neural networks via information,''
\newblock in {\em Why \& {When} {Deep} {Learning} {Works}: {Looking} {Inside}
  {Deep} {Learning}}, R.~Ronen, Ed. The Intel Collaborative Research Institute
  for Computational Intelligence (ICRI-CI), 2017.

\bibitem{Andrew}
A.~M. Saxe, Y.~Bansal, J.~Dapello, M.~Advani, A.~Kolchinsky, B.~D. Tracey, and
  D.~D. Cox,
\newblock ``On the information bottleneck theory of deep learning,''
\newblock in {\em Proc. {ICLR}}, 2018.

\bibitem{strouse_deterministic_2016}
D.~J. Strouse and D.~J. Schwab,
\newblock ``The deterministic information bottleneck,''
\newblock in {\em Proc. Uncertainty in Artificial Intelligence}, 2016.

\bibitem{kolchinsky_caveats_2018}
A.~Kolchinsky, B.~D. Tracey, and S.~V. Kuyk,
\newblock ``Caveats for information bottleneck in deterministic scenarios,''
\newblock in {\em Proc. ICLR}, 2018.

\bibitem{goldfeld_2019}
Z.~Goldfeld, E.~V.~D. Berg, K.~Greenewald, I.~Melnyk, N.~Nguyen, B.~Kingsbury,
  and Y.~Polyanskiy,
\newblock ``Estimating information flow in deep neural networks,''
\newblock in {\em Proc. {ICML}}, 2019, pp. 2299--2308.

\bibitem{Kolchinsky1}
A.~Kolchinsky and B.~D. Tracey,
\newblock ``Estimating mixture entropy with pairwise distances,''
\newblock {\em Entropy}, vol. 19, no. 7, pp. 361--377, 2017.

\bibitem{belghazi_mutual_2018}
M.~I. Belghazi, A.~Baratin, S.~Rajeshwar, S.~Ozair, Y.~Bengio, A.~Courville,
  and D.~Hjelm,
\newblock ``Mutual information neural estimation,''
\newblock in {\em Proc. ICML}, 2018, pp. 531--540.

\bibitem{Juan}
J.~C. Caicedo, A.~Goodman, K.~W. Karhohs, B.~A. Cimini, J.~Ackerman,
  M.~Haghighi, C.~Heng, T.~Becker, M.~Doan, C.~McQuin, M.~Rohban, S.~Singh, and
  A.~E. Carpenter,
\newblock ``Nucleus segmentation across imaging experiments: The 2018 data
  science bowl,''
\newblock {\em Nature Methods}, vol. 16, no. 12, pp. 1247--1253, 2019.

\bibitem{duda_et_al_2000}
R.~O. Duda, P.~E. Hart, and D.~G. Stork,
\newblock {\em Pattern Classification},
\newblock Wiley, 2nd edition, 2000.

\end{thebibliography}

\end{document}